\documentclass{article}

\usepackage[preprint, nonanonymous]{neurips_2026}

\usepackage[utf8]{inputenc}
\usepackage[T1]{fontenc}
\usepackage{hyperref}
\usepackage{url}
\usepackage{booktabs}
\usepackage{amsfonts}
\usepackage{amsmath}
\usepackage{nicefrac}
\usepackage{microtype}
\usepackage{xcolor}
\usepackage{graphicx}
\usepackage{array}
\usepackage{caption}
\usepackage{cleveref}
\usepackage{caption}
\usepackage{subcaption}
\usepackage{listings}
\usepackage{tabularx}
\usepackage{ragged2e}
\usepackage{diagbox}
\usepackage{bm}
\usepackage{tikz}
\usetikzlibrary{arrows.meta,fit,positioning}

\crefname{appendix}{Appendix}{Appendices}
\Crefname{appendix}{Appendix}{Appendices}

\graphicspath{{figs/}}

\title{Search, Fail, Recover: A Training Framework for Correction-Aware Reasoning}

\author{%
  \textbf{Dmitry Beresnev\textsuperscript{1,3} \quad
 Vladimir Makharev\textsuperscript{1,2}}\quad
  Roman Khalikov\textsuperscript{3}\\
  \textbf{Ivan Oseledets\textsuperscript{2}}\quad
  \textbf{Petr Anokhin\textsuperscript{2,3}}\\[0.5em]
  \normalfont\small
  \textsuperscript{1}Innopolis University, Innopolis, Russia \quad
  \textsuperscript{2}AXXX, Moscow, Russia\\
  \textsuperscript{3}Lomonosov Moscow State University, Moscow, Russia \\
  \texttt{d.beresnev.work@gmail.com}
}

\begin{document}

\maketitle

\begin{abstract}
    Many reasoning tasks are not well described by a single left-to-right chain: a solver may need to pursue a plausible branch, observe delayed failure, and return to the latest prefix that can still be completed. We introduce Pyligent, a training and inference framework inspired by the Diligent Learner formulation that represents reasoning as validated search over partial solution chains. A task validator labels generated continuations and failures, and the resulting search trees are converted into supervised targets for three actions: continue, finish, and backtrack, with optional traces that summarize abandoned branches. We evaluate Pyligent on a hidden directed graph task designed to isolate delayed-failure recovery, and on structured reasoning domains with exact validators, including $4{\times}4$ Sudoku, Sudoku with reasoning traces, and Blocksworld. Compared with gold-only supervised fine-tuning, Pyligent improves solve rate by $72.7$ percentage points on hidden graphs, by $17$ and $18$ points on mixed and expert Sudoku, by $27$ and $14$ points on mixed and expert Sudoku with reasoning traces, and by $13$ points on Blocksworld. These results suggest that explicit failed-branch supervision can teach useful recovery behavior beyond imitation of polished solution chains.
\end{abstract}

\section{Introduction}\label{sec:intro}

Reasoning is often presented as a linear chain of intermediate steps, but many hard tasks are not naturally linear. A solver may need to try a plausible direction, discover only later that the branch cannot lead to a solution, and return to the decision point where another continuation is possible. Standard chain-of-thought supervision hides this process: training data usually contains the polished successful path, not the failed attempts and recoveries that produced it.

\citep{shalev-shwartz_reasoning_2025} formalize this view in the Diligent Learner method. Their theory separates the ability to \emph{search} for a solution from the ability to \emph{validate} a proposed reasoning chain, and argues that validation is often much easier than generation. The central sufficient conditions are correspondingly behavioral rather than token-level: a learner should generate a correct next reasoning step with fixed nonzero probability, and when a branch is revealed to be wrong, it should learn to backtrack to the latest prefix that can still be completed. Their constructive proof shows that, under these assumptions, a depth-first search procedure with learned backtracking can avoid the exponential blowups that affect naive search, Tree-of-Thoughts-style branching, and pure gold-path imitation.

This paper studies the practical question left open by that theory: can we turn these conditions into an implementable training pipeline for contemporary, relatively small LLMs? We introduce Pyligent, a framework that represents reasoning as a tree of partial chains, uses a task validator to label generated continuations, and converts both successful and failed branches into supervised training examples. The resulting action space contains ordinary continuation and termination actions, plus explicit \texttt{<backtrack>} actions for recovery.

Pyligent instantiates the validator assumed by the theory with task-specific checkers, use it to label both successful continuations and failed leaves, and train models on three corresponding action types: continue, finish, and backtrack. The resulting experiments operationalize the Diligent Learner conditions in contemporary LLM training, and our backtrack categories measure whether generated recoveries return to repairable prefixes rather than merely matching the syntax of a recovery action.

Our most controlled test is the hidden directed graph task. The model only observes outgoing moves from the current node, so a locally legal action may later reveal a dead end or a failure state. Solving the task therefore requires trying next actions, identifying mistakes from delayed feedback, and returning to the relevant earlier choice point. We use this task to measure whether recovery can be learned directly, and then study how Pyligent transfers to more structured reasoning domains such as Sudoku and Blocksworld.

\section{Related work}\label{sec:related-work}
Chain-of-thought (CoT) prompting \citep{wei_chain--thought_2022} showed that exposing intermediate reasoning steps can substantially improve LLM performance on arithmetic, commonsense, and symbolic tasks. Most CoT supervision, however, presents only the successful trajectory: a solution is generated left-to-right, and failed branches are usually absent from the training signal. This makes CoT a useful representation of reasoning, but not by itself a training procedure for learning when to abandon an unproductive branch and resume from an earlier state.

Several lines of work address this limitation by adding search, critique, or revision around the model. Tree-of-Thoughts \citep{yao_tree_2023} and MCTS-based methods such as Reasoning-via-Planning \citep{hao_reasoning_2023} and AlphaLLM \citep{tian_toward_2024} externalize exploration into an explicit inference-time tree, using model- or discriminator-based evaluations to select promising continuations. Prompting-level refinement methods, including Self-Refine \citep{madaan_self-refine_2023}, Reflexion \citep{shinn_reflexion_2023}, and CRITIC \citep{gou_critic_2023}, instead keep the interaction in context by alternating generation with feedback or critique. These methods demonstrate that reasoning often benefits from non-linear exploration, especially when reliable feedback is available \citep{huang_large_2023,kamoi_when_2024}.

A complementary direction trains models directly on search-like traces. Stream of Search \citep{gandhi_stream_2024} serializes DFS and BFS trajectories; Searchformer \citep{lehnert_beyond_2024} and DualFormer \citep{su_dualformer_2024} distill planner traces into transformer models; Self-Backtracking \citep{yang_step_2025} introduces explicit \texttt{<backtrack>} actions; and ASTRO \citep{kim_astro_2025} turns MCTS rollouts into reasoning traces that include reflection and recovery. Larger-scale correction policies have also been trained with reinforcement learning, including SCoRe \citep{kumar_training_2024}, RISE \citep{qu_recursive_2024}, DeepSeek-R1 \citep{deepseek-ai_deepseek-r1_2025}, and Kimi k1.5 \citep{team_kimi_2025}. Our work follows this broad search-training view, but focuses on a specific source of supervision: failed branches produced by a task validator and converted into explicit recovery examples.

Recent empirical studies help explain why this supervision matters. \citep{gandhi_cognitive_2025} find that the structure of search traces can be more important than final-answer labels for eliciting verification and backtracking behaviors. At the same time, backtracking is not universally helpful in every domain or at every frequency: \citep{qin_backtrack_2025} and \citep{cai_how_2025} show that the usefulness of backtracking depends strongly on task structure and training distribution. These observations motivate our ablations over trace use, exploration policy, and failed-branch weighting.

The closest conceptual foundation for our work is the Diligent Learner method of \citep{shalev-shwartz_reasoning_2025}. Their theory separates generation from validation and gives sufficient conditions for CoT learnability in terms of two behavioral capabilities: the model should generate a valid next semantic step with nonzero probability, and when a branch is found to be incorrect, it should backtrack to $\beta(c)$, the maximal prefix that can still be completed. Their analysis further shows why the target of a backtrack matters: returning only to the immediate parent can be insufficient when a whole sub-branch is invalid, while overshooting can discard useful progress.


\section{Framework methodology}\label{sec:framework-methodology}

\begin{figure}[h]
    \centering
    \includegraphics[width=0.85\linewidth]{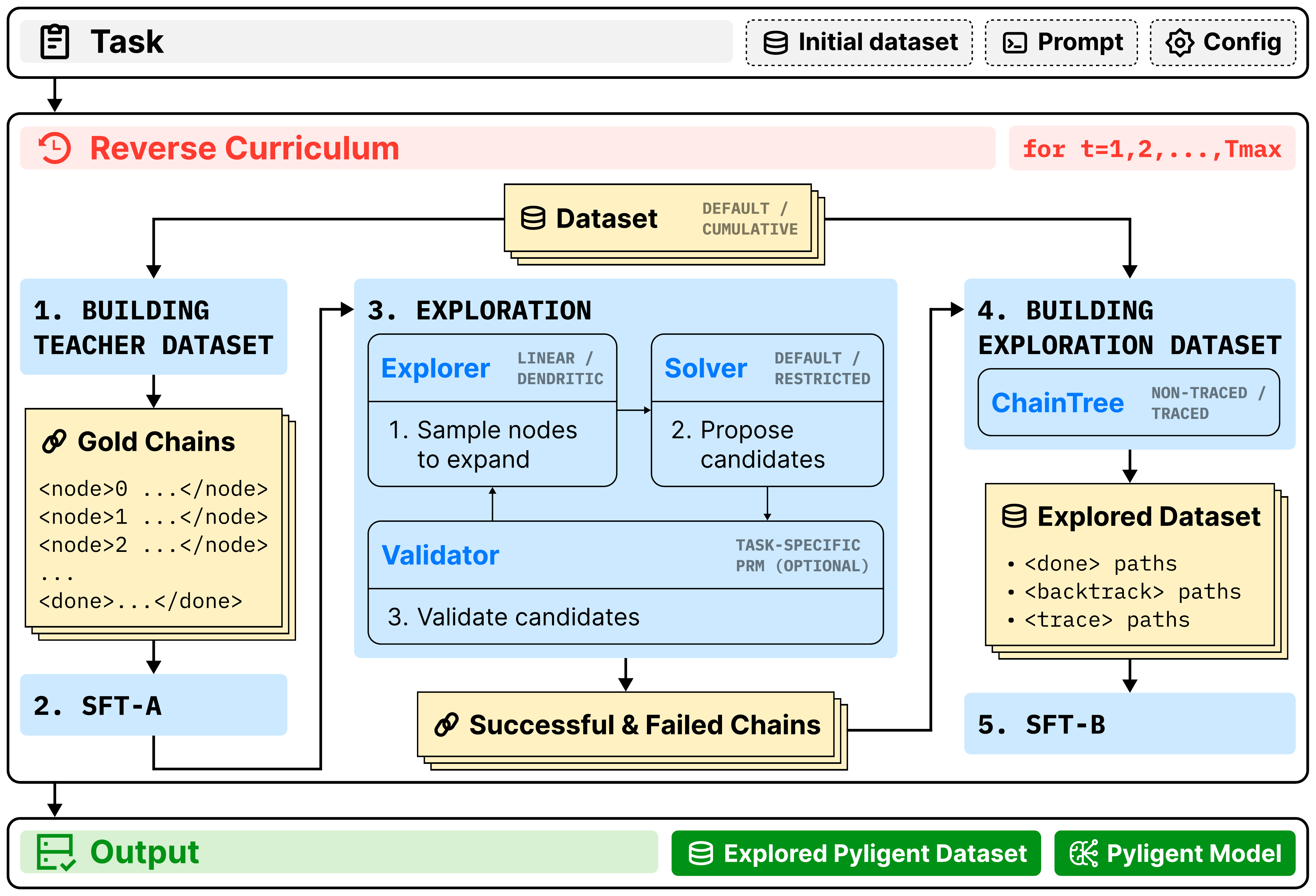}
    \caption{The Pyligent framework. The pipeline iterates
        for $t = 1, \ldots, T_{\max}$ rounds, alternating between fine-tuning on
        teacher-generated gold chains~(SFT-A) and exploration-driven fine-tuning
        on \emph{ChainTree}-structured trajectories containing successful, backtracked,
        and traced paths~(SFT-B).}
    \label{fig:methodology}
\end{figure}

Pyligent turns chain-of-thought generation into a validated search over partial solution chains (\Cref{fig:methodology}). Instead of training only on polished gold trajectories, it records successful continuations and failed branches, then converts them into supervised examples for continuing, terminating, and recovering. In the language of \citep{shalev-shwartz_reasoning_2025}, the solver supplies approximate next-step generation and the task validator supplies retrospective correctness labels.

\subsection{Representation and components}\label{subsec:search-representation}

For each problem instance, we maintain a tree of partial reasoning states. A state is a prefix
\[
    c_t = (a_0, a_1, \ldots, a_t),
\]
where each action uses a restricted markup vocabulary:
\begin{small}
    \[
        \texttt{<node>ID CONTENT</node>}, \quad
        \texttt{<done>ANSWER</done>}, \quad
        \texttt{<backtrack>ID REASON</backtrack>}.
    \]
\end{small}

Here \texttt{<node>} appends one intermediate step, \texttt{<done>} proposes a final answer, and \texttt{<backtrack>} returns control to an earlier node. The explicit node identifiers make branch structure visible to both the data-generation pipeline and the inference runtime.

The framework has four components. The \emph{solver} is the trainable language model, which emits one action at a time. The \emph{explorer} selects prefixes from which the solver should sample continuations and controls how much breadth or depth to allocate. The \emph{validator} checks whether each proposed action is legal for the task state. The \emph{ChainTree} stores accepted and rejected branches, assigns backtrack targets for failed branches, and converts the resulting search tree into prefix-to-action training examples. We use linear and dendritic explorers in the experiments; the former keeps cost predictable by extending a bounded number of parallel chains, while the latter starts broad and then concentrates budget on deeper continuations. Full explorer definitions are given in \Cref{apx:explorer-details}.

\subsection{Training pipeline}\label{subsec:training-pipeline}

Training proceeds in three stages. First, SFT-A trains the solver on gold solution chains, teaching the action format and giving the model a nonzero chance of proposing valid next steps. Second, validator-guided exploration runs the solver from selected prefixes and records both valid continuations and failed branches. Third, SFT-B fine-tunes on the examples constructed from the resulting ChainTree.

SFT-B contains success pairs, failure pairs, and, for traced recovery, continuation-after-trace pairs. Success pairs train the model to extend valid generated or gold branches. Failure pairs train the model to emit a \texttt{<backtrack>} action when a branch violates task constraints. For a failed branch that diverges after node $i$ and reaches invalid leaf $j$, ChainTree creates the target
\[
    c_j \mapsto \texttt{<backtrack>}i~r\texttt{</backtrack>},
\]
where $r$ is a validator-derived reason. We use a reverse curriculum over prefixes: exploration starts near the end of a gold chain, where the remaining continuation is short, and later moves toward the root so that the solver must handle longer recovery-aware trajectories. Pair construction and curriculum details are shown in \Cref{apx:sftb-pairs}.

\subsection{Inference and traced recovery}\label{subsec:inference}

At inference time, the solver generates one action at a time. A \texttt{<node>} action is appended to the current chain, a \texttt{<done>} action terminates with the proposed answer, and a \texttt{<backtrack>} action is checked for a valid target before the runtime truncates the context to that node. Inference stops when the solver emits \texttt{<done>} or exhausts a fixed step budget.

Plain backtracking removes the failed branch from context, which can lead the solver to repeat the same mistake. To preserve compact failure information, traced recovery inserts a runtime-generated \texttt{<trace>} after truncation. The trace records the validator reason and a short summary of the abandoned branch; it is never a model target. During SFT-B we inject synthetic traces for matching failure examples, producing continuation pairs that teach the model how to resume after recovery. The exact trace format is given in \Cref{apx:traces}.

\section{Tasks methodology}\label{sec:tasks-methodology}
\subsection{Hidden directed graph task}\label{subsec:graph-hidden-task}

The hidden directed graph task is a controlled benchmark for testing whether a solver has learned an effective search and backtrack policy. Each instance is a directed graph with distinguished nodes \texttt{START}, \texttt{GOAL}, and \texttt{FAIL}. The model must find a path from \texttt{START} to \texttt{GOAL} while avoiding \texttt{FAIL}, but it only observes the outgoing moves from its current node. After the model chooses a legal non-terminal move, the environment reveals the outgoing moves from the newly reached node as the next observation. This makes the task useful for evaluating Pyligent under the Diligent Learner assumptions: the model cannot solve by copying a static problem description, and a locally valid move can still expose a branch that must be abandoned.

This task is intentionally close to the search-theoretic examples in \citep{shalev-shwartz_reasoning_2025}. The difficulty is not parsing the local action format: from any visible node, choosing an outgoing edge is easy to validate. The difficulty is that a locally valid edge can be globally wrong, and the evidence for that mistake may appear only several moves later. A successful solver must therefore learn to propose plausible moves under partial information and, once a dead end or \texttt{FAIL} node is revealed, return to the latest useful choice point.

Formally, an instance is a directed graph $G=(V,E)$ with start node $s$, goal node $g$, and forbidden failure node $d$. At step $t$, the environment state is the explored path $p_t=(s,v_1,\ldots,v_t)$, but the model-facing observation exposes only the outgoing neighbors $O(v_t)=\{u:(v_t,u)\in E\}$. The solver must choose a visible neighbor, terminate when the goal is visible, or backtrack when the current branch is invalid or revealed to be unsolvable.

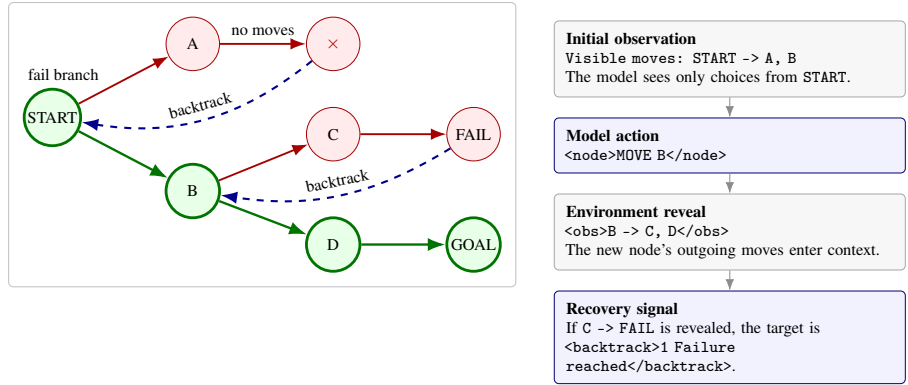
\begin{figure*}[t]
    \centering
    \resizebox{0.85\linewidth}{!}{%
        \begin{tikzpicture}[
                >=Latex,
                graphnode/.style={circle, draw=black!55, fill=white, minimum size=8mm, inner sep=1pt, font=\scriptsize},
                goodnode/.style={graphnode, draw=green!45!black, fill=green!9, very thick},
                badnode/.style={graphnode, draw=red!65!black, fill=red!8},
                goodedge/.style={->, line width=1pt, draw=green!45!black},
                badedge/.style={->, line width=0.8pt, draw=red!65!black},
                backtrackedge/.style={->, dashed, line width=0.9pt, draw=blue!55!black},
                panel/.style={draw=black!35, fill=black!3, rounded corners=2pt, inner sep=5pt, font=\scriptsize, align=left, text width=4.9cm},
                actionpanel/.style={draw=blue!45!black, fill=blue!5, rounded corners=2pt, inner sep=5pt, font=\scriptsize, align=left, text width=4.9cm}
            ]
            \node[goodnode] (start) at (0,0) {START};
            \node[badnode] (a) at (2.1,1.1) {A};
            \node[goodnode] (b) at (2.1,-1.1) {B};
            \node[badnode] (empty) at (4.2,1.1) {\textcolor{red!70!black}{$\times$}};
            \node[badnode] (c) at (4.2,-0.25) {C};
            \node[goodnode] (d) at (4.2,-1.9) {D};
            \node[badnode] (fail) at (6.3,-0.25) {FAIL};
            \node[goodnode] (goal) at (6.3,-1.9) {GOAL};

            \draw[badedge] (start) -- node[pos=0.32, above left, font=\scriptsize] {fail branch} (a);
            \draw[goodedge] (start) -- (b);
            \draw[badedge] (a) -- node[above, font=\scriptsize] {no moves} (empty);
            \draw[badedge] (b) -- (c);
            \draw[goodedge] (b) -- (d);
            \draw[badedge] (c) -- (fail);
            \draw[goodedge] (d) -- (goal);
            \draw[backtrackedge] (empty) to[bend left=24] node[above, sloped, font=\scriptsize] {backtrack} (start);
            \draw[backtrackedge] (fail) to[bend left=20] node[above, sloped, font=\scriptsize] {backtrack} (b);
            \node[draw=black!25, rounded corners=2pt, fit=(start)(a)(b)(empty)(c)(d)(fail)(goal), inner sep=6pt,
            ] {};

            \node[panel, anchor=north west] (ctx0) at (7.5,1.45)
            {\textbf{Initial observation}\\
                \texttt{Visible moves: START -> A, B}\\
                The model sees only choices from \texttt{START}.};
            \node[actionpanel, below=3mm of ctx0] (move)
            {\textbf{Model action}\\
                \texttt{<node>MOVE B</node>}};
            \node[panel, below=3mm of move] (ctx1)
            {\textbf{Environment reveal}\\
                \texttt{<obs>B -> C, D</obs>}\\
                The new node's outgoing moves enter context.};
            \node[actionpanel, below=3mm of ctx1] (bt)
            {\textbf{Recovery signal}\\
                If \texttt{C -> FAIL} is revealed, the target is\\
                \texttt{<backtrack>1 Failure reached</backtrack>}.};

            \draw[->, draw=black!45] (ctx0) -- (move);
            \draw[->, draw=black!45] (move) -- (ctx1);
            \draw[->, draw=black!45] (ctx1) -- (bt);
        \end{tikzpicture}%
    }
    \caption{Hidden directed graph interaction. The environment contains a full graph, but the prompt exposes only the current node's outgoing moves. Accepted moves reveal a new \texttt{<obs>} line; branches that reveal \texttt{FAIL} or no outgoing moves become supervised backtrack examples.}\label{fig:graph-hidden-task}
\end{figure*}

The model-facing action space is deliberately small, so task performance mostly reflects search decisions rather than output-format complexity. A non-terminal move is written as \texttt{<node>MOVE X</node>}, a terminal answer as \texttt{<done>START -> ... -> GOAL</done>}, and recovery as \texttt{<backtrack>ID REASON</backtrack>}. Accepted moves reveal a new observation, rendered separately as \texttt{<obs>...</obs>}, so the target action remains the decision rather than asking the model to hallucinate hidden graph structure.

The validator is exact and enforces local visibility, terminal-action consistency, and recovery from revealed dead ends. Thus the task separates two questions cleanly: whether the model can choose plausible visible moves, and whether it can return to the appropriate earlier choice point after a locally valid branch becomes impossible. All hidden directed graph experiments use Qwen3-0.6B and 300 training examples; full validation rules are given in \Cref{apx:task-methodology-details}.

\subsection{Sudoku 4x4 task}\label{subsec:sudoku-task}

We instantiate Pyligent on $4{\times}4$ Sudoku. The grid contains digits $1$--$4$, empty cells are rendered as \texttt{\_}, and the four $2{\times}2$ boxes are separated in the text representation. Each row, column, and box must contain each digit exactly once in a completed solution. The task is small enough to permit exact validation, but still exposes the failure mode targeted by the Diligent Learner theory: a locally plausible fill can make a later branch invalid unless the model can return to an earlier state.

\subsubsection{Gold chains and validation}\label{subsubsec:sudoku-validation}

Each Sudoku action contains a complete grid rather than a coordinate-level edit, and the solver emits exactly one \texttt{<node>}, \texttt{<done>}, or \texttt{<backtrack>} tag per turn. The full action format is given in \Cref{apx:task-methodology-details}.

Gold chains are stepwise grid completions. \Cref{fig:sudoku} (green arrows) shows a representative path: the initial puzzle is node 0 and each subsequent node fills exactly one cell. The terminal \texttt{<done>} action must repeat the completed grid. Invalid generated branches are converted into backtrack targets rather than being discarded (\Cref{fig:sudoku}, red arrows); their validator reasons become the natural-language reason field in backtrack targets and, in traced mode, the reason field inside the injected trace. \Cref{apx:task-methodology-details} gives the exact Sudoku validation rules.

\subsubsection{Experimental setup}\label{subsubsec:sudoku-setup}

We evaluate two $4{\times}4$ Sudoku datasets. The \emph{mixed} train dataset contains 1,200 training chains: 300 examples for each of the easy, medium, hard, and expert difficulty buckets. Difficulty is controlled through the number of pre-filled cells; the generation configuration uses minimum clue counts of 14, 10, 8, and 6 respectively for the four buckets. Evaluation \emph{mixed} dataset contains 50 puzzles for each difficulty (200 puzzles in total). The \emph{expert} datasets contains only expert-difficulty puzzles --- 1,200 for training and 200 for evaluation, --- using the same minimum clue count of 6. All Sudoku experiments use Qwen3-4B with QLoRA fine-tuning.

\subsection{Sudoku w/ RT 4x4 Task}\label{subsec:sudoku-extended-task}

The Sudoku with reasoning trances (w/ RT) task augments standard Sudoku solving by attaching a natural-language explanation to each grid transition. The motivation is practical rather than evaluative: when a model spells out \emph{why} a particular cell is filled, each step becomes easier to inspect and recover from. This variant tests whether the same validated-search pipeline remains useful when the action contains both a verifiable grid transition and an unverified natural-language rationale.

Each chain begins with a partially filled Sudoku grid and each \texttt{<node>} action contains a grid followed by a brief justification. The dataset is produced by sampling Sudoku puzzles, solving them into one-cell-at-a-time reference chains, and prompting Qwen3-32B to add concise explanations for each transition. Grid states are validated exactly as in the standard Sudoku task, while the natural-language reasoning is not validated. Further generation details and the prompt template are given in \Cref{apx:task-methodology-details,apx:sudoku-extended-generation-prompt}.

\subsection{Blocksworld task}\label{subsec:blocksworld-task}

Blocksworld is a classical planning domain in which a set of lettered blocks must be rearranged from an initial stack configuration into a specified goal configuration using four primitive actions: pick up, put down, stack, and unstack. Subject to strict physical preconditions: only one block may be held at a time, and a block can only be picked up or unstacked if nothing is on top of it \citep{valmeekam_planbench_2023}. Despite its apparent simplicity, Blocksworld has become a canonical stress-test for LLM planning: standard autoregressive models and even chain-of-thought prompting fail reliably on instances beyond a few blocks, because valid plans require maintaining a consistent world-state across many interdependent steps and recovering gracefully when a chosen action renders a necessary precondition unreachable \citep{valmeekam_large_2022,kambhampati_position_2024}.
\paragraph{Data generation}\label{subsubsec:blocksworld-data-gen}
We generate Blocksworld instances with 3 to 12 blocks using the standard \citep{valmeekam_planbench_2023} framework, solve valid instances with the Fast-Forward planner \citep{hoffmann_ff_2001}, and convert the resulting PDDL plans into text action chains. In total, we generated 9{,}132 tasks with plan lengths ranging from 3 to 39 actions, reserving 30\% for testing while preserving the plan-length distribution. To decrease the amount of linearly increasing solutions during exploration, we preserved tasks with up to 16 actions required to solve them. The final size of the training dataset was 3 {,}035 tasks. As a test set, we used full plan-length distribution with up to 42 actions and total size of 2{,}740 tasks. Additional generation details are given in \Cref{apx:task-methodology-details}.
\paragraph{Validation}\label{subsubsec:blocksworld-val}
The correctness is determined by VAL \citep{howey_val_2004}, which symbolically simulates the candidate plan and checks both the action preconditions and the satisfaction of the final goal. This provides an exact validator for the planning domain, analogous to the graph and Sudoku validators.

\section{Results}\label{sec:results}
This section presents the evaluation protocol used to analyze correction behavior. Unless stated otherwise, aggregate results are reported as mean $\pm$ standard deviation over three runs. The primary task metric is percent of solved examples: a chain succeeds when its terminal \texttt{<done>} action parses as a complete valid answer for the given problem. We also track search cost through generated steps, generated length, and the number of backtracks. Because Pyligent explicitly trains recovery, we additionally classify every emitted backtrack into four categories: invalid, valid, correct and perfect (\Cref{tab:backtrack-categories}).


We report two checkpoint families. A \emph{Pyligent checkpoint} is obtained by the full iterative pipeline: train, explore, convert the resulting search trees into supervision, and repeat for $T$ iterations. A \emph{finetuned checkpoint}, referred in tables and plots usually as \emph{model FT}, is a fresh copy of the base model fine-tuned once on the final dataset collected by the Pyligent pipeline. Thus, Pyligent checkpoints see the data through the iterative training-and-exploration loop, while model FT checkpoints are trained on the full Pyligent dataset at once. We compare both families against the base model and a gold-only fine-tuned base model reference.

\subsection{Hidden directed graph}\label{subsec:hidden-directed-graph-results}

\begin{table*}[t]
    \centering
    \small
    \setlength{\tabcolsep}{6pt}
    \caption{Hidden directed graph evaluation results, reported as mean $\pm$ standard deviation. The final column reports the percentage of all generated backtracks that are perfect; the full mutually exclusive category breakdown is shown in \Cref{fig:graph-hidden-bt}.}\label{tab:graph-hidden-results}
    \begin{tabular*}{\linewidth}{@{\extracolsep{\fill}}lcccc@{}}
        \toprule
        Checkpoint & Success (\%) & Avg. steps & Avg. backtracks & Perfect BT (\%) \\
        \midrule
        Base model              & $0.0 \pm 0.0$  & $29.85 \pm 0.05$ & $0.00 \pm 0.00$ & --     \\
        Base model + thinking   & $0.2 \pm 0.3$  & $14.69 \pm 0.26$ & $0.00 \pm 0.00$ & --     \\
        Gold-only SFT           & $3.8 \pm 1.2$  & $24.21 \pm 1.22$ & $0.47 \pm 0.03$ & $23.2$ \\
        Pyligent                & $76.5 \pm 2.3$ & $10.96 \pm 0.43$ & $2.14 \pm 0.16$ & $47.7$ \\
        \bottomrule
    \end{tabular*}
\end{table*}

\begin{figure*}[t]
    \centering
    \includegraphics[width=.82\linewidth]{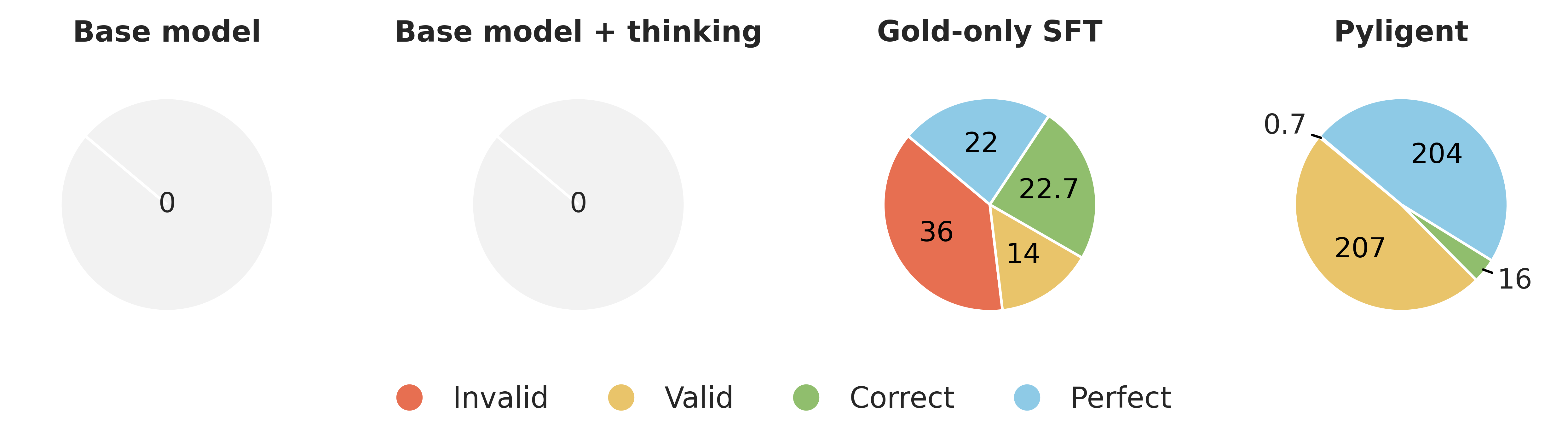}
    \caption{Hidden directed graph backtrack quality. Pie charts show mean backtrack counts over three runs, using the mutually exclusive invalid, valid-only, correct-only, and perfect categories from \Cref{tab:backtrack-categories}.}\label{fig:graph-hidden-bt}
\end{figure*}

The hidden graph results show a sharp separation between imitating successful paths and learning recovery behavior. The base model without thinking never solves an example, while thinking gives only $0.2 \pm 0.3\%$ success and neither setting emits usable backtracks (\Cref{tab:graph-hidden-results} and \Cref{fig:graph-hidden-bt}). Gold-only SFT gives a small improvement to $3.8 \pm 1.2\%$ success, but it still behaves mostly like a long forward search: average generated length is $24.21 \pm 1.22$ actions against an average gold length of $5.0$, and only $23.2\%$ of its backtracks are perfect.

By contrast, the Pyligent checkpoint reaches $76.5 \pm 2.3\%$ success with much shorter generations and active correction. Its average generated length is $10.96 \pm 0.43$ actions, less than half the gold-only SFT length, and it emits $427.7 \pm 31.1$ backtracks across 200 examples. Almost all of these are non-invalid, with $47.7\%$ perfect backtracks, meaning the model often returns to the immediate choice point responsible for the failed branch rather than merely jumping to some earlier valid node. At the example level, $11.0 \pm 5.8\%$ of runs containing at least one valid-only backtrack still finish successfully, and each such run later emits a correct or perfect backtrack that repairs the earlier imprecise recovery. The matched comparison is therefore consistent with the goal of the task: validator-guided failed branches provide the signal needed to learn when to abandon a revealed dead end, while gold-only training mostly teaches the format of successful paths.

\subsection{Sudoku 4x4}\label{subsec:sudoku-results}

\begin{table}[ht]
    \centering
    \footnotesize
    \captionsetup{skip=10pt}
    \begingroup
    \setlength{\tabcolsep}{7pt}
    \renewcommand{\arraystretch}{1.45}
    \begin{tabular}{@{}l|cccccc@{}}
        \diagbox[width=3.2cm,height=0.7cm,outerleftsep=0.1cm]{\strut Task}{\strut Model}
                                           & Linear          & Dendritic  & Linear FT  & Dendritic FT & Gold-only SFT & Base       \\ \hline
        Sudoku $4{\times}4$ (mixed)        & $\bm{82\pm1.0}$ & $81\pm0.3$ & $69\pm1.0$ & $74\pm1.8$   & $65\pm0.3$    & $11\pm0.9$ \\
        Sudoku $4{\times}4$ (expert)       & $\bm{66\pm0.8}$ & $54\pm0.3$ & $56\pm1.6$ & $55\pm1.2$   & $48\pm1.0$    & $0\pm0.0$  \\
        \hline
        Sudoku w/ RT $4{\times}4$ (mixed)  & \bm{$58\pm1.3$} & ---        & ---        & ---          & $31\pm1.1$    & $8\pm0.8$  \\
        Sudoku w/ RT $4{\times}4$ (expert) & \bm{$34\pm1.0$} & ---        & ---        & ---          & $20\pm0.8$    & $0\pm0.0$  \\
    \end{tabular}
    \endgroup
    \caption{Sudoku $4{\times}4$ results for Qwen3-4B. All entries are solved validation puzzles (\%) reported as mean $\pm$ standard deviation over three runs. Pyligent architectures use traces and one failed-path epoch. `Sudoku w/ RT' refers to Sudoku $4{\times}4$ with reasoning traces.}\label{tab:sudoku-results-summary}
\end{table}

\begin{figure}[ht]
    \centering
    \includegraphics[width=\linewidth]{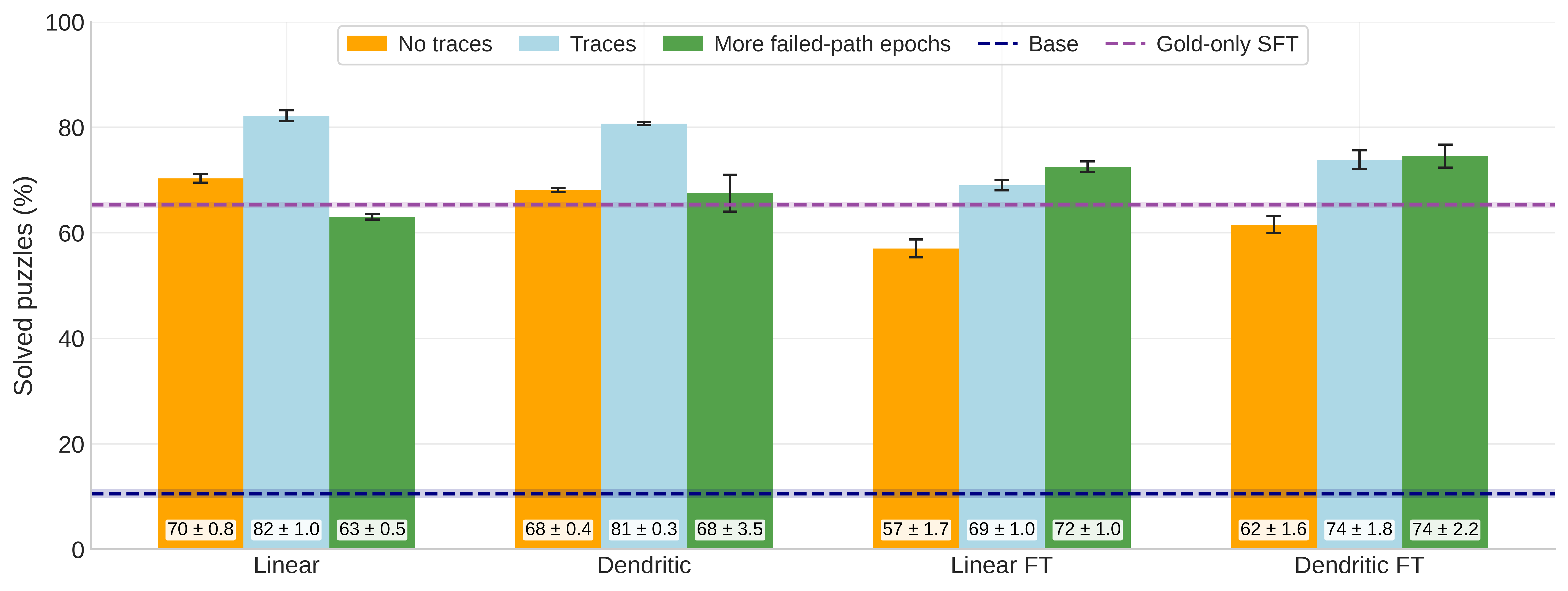}
    \caption{Sudoku $4{\times}4$ ablations on the mixed dataset with Qwen3-4B, reported as mean $\pm$ standard deviation over three runs. `No traces' uses architectures without traces and one failed-path epoch. `Traces' uses architectures with traces and one failed-path epoch, matching the first-row setting in \Cref{tab:sudoku-results-summary}. `More failed-path epochs' uses traces and three failed-path epochs.}\label{fig:sudoku-results}
\end{figure}

\begin{figure}[ht]
    \centering
    \includegraphics[width=.75\linewidth]{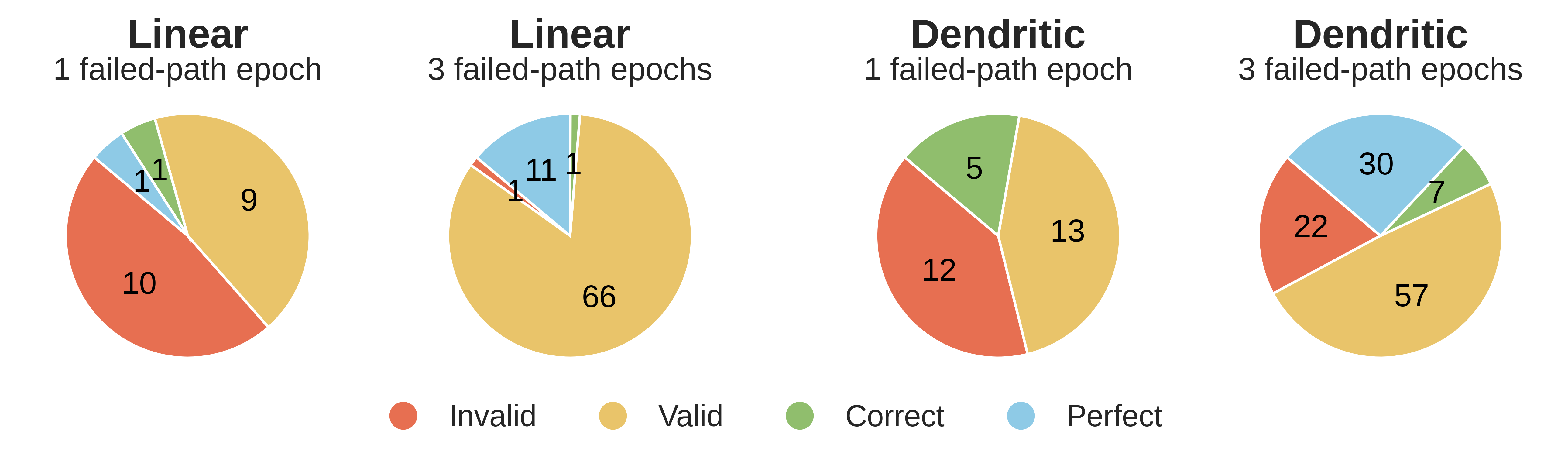}
    \caption{Sudoku $4{\times}4$ backtrack quality on the mixed dataset with Qwen3-4B. Pie charts show mean backtrack counts over three runs. All listed architectures use traces.}\label{fig:sudoku-bt}
\end{figure}

\subsubsection{Sudoku 4x4 Task}

The Sudoku results show that Pyligent models improve over both gold-only SFT and the base model on both datasets. On the mixed validation split, the Pyligent model with linear explorer reaches $82 \pm 1.0\%$, compared with $65 \pm 0.3\%$ for gold-only SFT and $11 \pm 0.9\%$ for the base model (\Cref{tab:sudoku-results-summary}). On the expert-only split, absolute accuracy is lower for every model, as expected from the harder puzzle distribution, but the gap remains large: the best Pyligent model reaches $66 \pm 0.8\%$, compared with $48 \pm 1.0\%$ for gold-only SFT and $0 \pm 0.0\%$ for the base model. The finetuned checkpoints also improve over the base model, but they do not match the best Pyligent checkpoints, indicating that the iterative training-and-exploration loop is useful beyond just collecting a larger supervised dataset.

The ablations demonstrate two practical effects on the mixed dataset. First, moving from no traces to traces notably improves performance across the Pyligent and model FT variants (\Cref{fig:sudoku-results}). This supports the motivation for traced recovery: preserving a compact account of the failed branch helps the model continue after backtracking instead of repeating the same local mistake. Second, increasing failed-path training from one epoch to three epochs usually reduces solve rate for Pyligent checkpoints, although finetuned checkpoints do not show the same drop. The backtrack-quality plots explain part of this tradeoff: the extra failed-path focus improves the quality of generated backtracks, but better backtrack syntax and target selection do not automatically translate into better final continuation (\Cref{fig:sudoku-bt}).

\subsubsection{Sudoku w/ RT 4x4}\label{subsec:sudoku-extended-results}

Using the same training set size as in the standard Sudoku $4{\times}4$ setting,
Sudoku $4{\times}4$ with reasoning traces (w/ RT) allows the model to generate
not only completed grids but also natural-language reasoning steps.
The goal is to give the model more room to `think' before making each prediction.

Although the overall scores are generally lower than for standard Sudoku $4{\times}4$,
the superiority of the Pyligent models is preserved (\Cref{tab:sudoku-results-summary}).
While reasoning traces appear to be more harmful than helpful in this particular setting,
the experiment nevertheless suggests that the proposed Pyligent framework may still
improve performance on tasks that require natural-language reasoning, such as MATH and GSM8K.







\subsection{Blocksworld}\label{subsec:blocksworld-results}

\begin{table*}[t]
    \centering
    \small
    \setlength{\tabcolsep}{6pt}
    \caption{Blocksworld evaluation results, reported as mean $\pm$ standard deviation. The final column reports the percentage of all generated backtracks that are perfect; the full mutually exclusive category breakdown is shown in \Cref{fig:blocksworld-bt}.}\label{tab:blocksworld-results}
    \begin{tabular*}{\linewidth}{@{\extracolsep{\fill}}lcccc@{}}
        \toprule
        Checkpoint & Success (\%) & Avg. steps & Avg. backtracks & Perfect BT (\%) \\
        \midrule
        Base model              & $0.0 \pm 0.0$  & $199.52 \pm 0.11$ & $113.85 \pm 0.35$ & $5.24$     \\
        Gold-only SFT           & $29.37 \pm 0.15$ & $47.99 \pm 6.31$ & $0.03 \pm 0.02$ & $2.9$ \\
        Pyligent                & $50.35 \pm 0.12$ & $46.99 \pm 0.57$ & $4.54 \pm 0.15$ & $14.4$ \\
        \bottomrule
    \end{tabular*}
\end{table*}

\begin{figure*}[t]
    \centering
    \includegraphics[width=.82\linewidth]{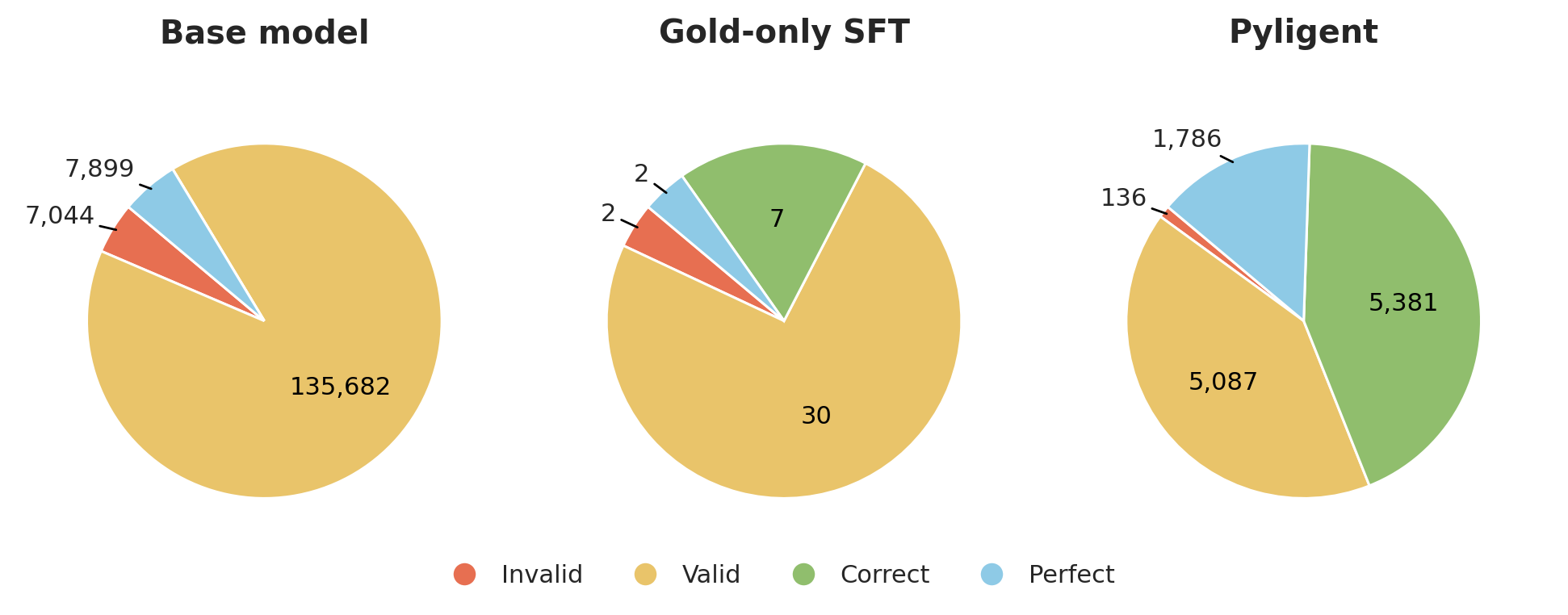}
    \caption{Blocksworld backtrack quality. Pie charts show mean backtrack counts over three runs, using the mutually exclusive invalid, valid-only, correct-only, and perfect categories from \Cref{tab:backtrack-categories}.}\label{fig:blocksworld-bt}
\end{figure*}

In Blocksworld, the base model is again unable to solve the task and emits a large amount of valid backtracks without any improvement (\Cref{tab:blocksworld-results} and \Cref{fig:blocksworld-bt}). Gold-only SFT improves the accuracy up to $29.37 \pm 0.15$ but the model failed to emit any reasonable number of backtracks. The Pyligent model solves $50.35 \pm 0.12\%$ of validation tasks and achieves more than 20\% accuracy against the gold-only SFT model. It learns how to perform backtracks, and additional training examples increased the capabilities of the model:  nearly 700 tasks were uniquely solved by Pyligent across 3 runs and backtracks helped to solve more than 11\% of these tasks (See \Cref{apx:blocksworld-analysis} for a task success distribution analysis).

Blocksworld is a much less forgiving domain than the Sudoku because local action validity does not imply global progress toward the goal. A generated plan may satisfy every precondition for many steps, while still dismantling useful structure or postponing a necessary subgoal. However, a proper fine-tuning and backtracks improved the Qwen3-4B capabilities noticeably.

\section{Conclusion}\label{sec:discussion}

The results should be read as an empirical implementation study of the Diligent Learner hypothesis. Pyligent turns the theory's behavioral conditions into a concrete training recipe for smaller LLMs by combining gold-chain SFT, validator-guided exploration, supervised backtrack targets, and traced continuation after recovery.

The hidden directed graph task provides the cleanest evidence for this framing. Gold-only SFT learns the surface format of successful paths but rarely recovers from mistakes, while the Pyligent checkpoint actively uses valid and often perfect backtracks to solve substantially more examples. This supports the claim that the useful signal is not only the final answer or the polished reasoning path, but the structure of failed attempts and the correct point of return. In this sense, the graph task is not just another benchmark; it is a direct probe of whether the model has learned the recoverability behavior required by the theory.

The broader tasks show both promise and limits. Sudoku and Blocksworld supply exact validators, making them natural domains for the same mechanism, but the ablations indicate that training recovery is sensitive to trace format, exploration policy, checkpoint family, and failure-data weighting. In particular, increasing exposure to failed branches can improve the number of non-invalid backtracks without reliably improving final solve rate. This suggests that learning to emit a syntactically valid recovery action is easier than learning the full policy of when to abandon a branch and where to resume. Future work should therefore separate next-action quality, failure detection, backtrack target quality, and post-backtrack continuation more explicitly.

Overall, our findings are consistent with the search-theoretic motivation while making its engineering requirements concrete. Reliable backtracking is not a behavior that should be expected to emerge automatically from gold solutions alone; it needs to be represented in the action space, generated during exploration, validated in hindsight, and trained as an explicit target.

\section{Limitations and Future Work}\label{sec:limitations-future-work}

Pyligent is currently evaluated on small or synthetic tasks with exact validators, so its behavior may not transfer directly to broader text-based reasoning settings where validation is weaker, delayed, or ambiguous. The present experiments also use relatively small solver models and focus on domains where the action format can be constrained tightly. It remains unclear how much of the observed recovery behavior would persist with open-ended natural-language actions.

The framework also adds cost. Exploration requires repeated solver sampling and validator calls; traced recovery increases context length; and failed-branch supervision introduces a weighting choice that can improve backtrack quality without improving final solve rate. Future work should measure this accuracy-cost tradeoff more systematically.

Finally, promising extensions include validators for rationale quality in reasoning-annotated tasks, larger solver models, and evaluation on natural-language benchmarks where verification can be decomposed into smaller checked steps.

\begin{small}
    \bibliographystyle{plainnat}
    \bibliography{references}
\end{small}

\clearpage
\section{Appendix}
\crefalias{subsection}{appendix}
\crefalias{subsubsection}{appendix}

\newcommand{\sudokuStateA}{\begingroup\setlength{\tabcolsep}{0pt}\renewcommand{\arraystretch}{0.88}\begin{tabular}{@{}c@{}}
        \texttt{3\_\,|\,12} \\
        \texttt{21\,|\,43}  \\[-0.4mm]
        \texttt{----+----}  \\[-0.4mm]
        \texttt{43\,|\,21}  \\
        \texttt{1\_\,|\,34}
    \end{tabular}\endgroup}
\newcommand{\sudokuStateB}{\begingroup\setlength{\tabcolsep}{0pt}\renewcommand{\arraystretch}{0.88}\begin{tabular}{@{}c@{}}
        \texttt{34\,|\,12}  \\
        \texttt{21\,|\,43}  \\[-0.4mm]
        \texttt{-----+----} \\[-0.4mm]
        \texttt{43\,|\,21}  \\
        \texttt{1\_\,|\,34}
    \end{tabular}\endgroup}
\newcommand{\sudokuStateC}{\begingroup\setlength{\tabcolsep}{0pt}\renewcommand{\arraystretch}{0.88}\begin{tabular}{@{}c@{}}
        \texttt{34\,|\,12} \\
        \texttt{21\,|\,43} \\[-0.4mm]
        \texttt{----+----} \\[-0.4mm]
        \texttt{43\,|\,21} \\
        \texttt{12\,|\,34}
    \end{tabular}\endgroup}
\newcommand{\sudokuStateBad}{\begingroup\setlength{\tabcolsep}{0pt}\renewcommand{\arraystretch}{0.88}\begin{tabular}{@{}c@{}}
        \texttt{34\,|\,12} \\
        \texttt{21\,|\,43} \\[-0.4mm]
        \texttt{----+----} \\[-0.4mm]
        \texttt{43\,|\,21} \\
        \texttt{13\,|\,34}
    \end{tabular}\endgroup}

\begin{figure*}[ht]
    \centering
    \resizebox{0.85\linewidth}{!}{%
        \begin{tikzpicture}[
                >=Latex,
                sudokunode/.style={draw=black!55, fill=blue!7, rounded corners=2pt, minimum width=13mm, minimum height=12mm, inner sep=3pt, font=\tiny},
                badnode/.style={draw=red!65!black, fill=red!8, rounded corners=2pt, minimum width=13mm, minimum height=12mm, inner sep=3pt, font=\tiny},
                goalnode/.style={draw=green!45!black, fill=green!9, rounded corners=2pt, very thick, minimum width=13mm, minimum height=12mm, inner sep=3pt, font=\tiny},
                goodedge/.style={->, line width=0.95pt, draw=green!45!black},
                badedge/.style={->, line width=0.85pt, draw=red!65!black},
                backtrackedge/.style={->, dashed, line width=0.9pt, draw=blue!55!black},
                faillabel/.style={draw=red!65!black, fill=red!6, rounded corners=2pt, inner sep=3pt, font=\scriptsize, align=center},
                donelabel/.style={draw=green!65!black, fill=green!6, rounded corners=2pt, inner sep=3pt, font=\scriptsize, align=center},
                steplabel/.style={font=\scriptsize, fill=white, inner sep=1pt}
            ]
            \node[sudokunode] (s0) at (0,0) {\sudokuStateA};
            \node[sudokunode] (s1) at (3.4,0) {\sudokuStateB};
            \node[badnode] (bad) at (6.8,1.15) {\sudokuStateBad};
            \node[goalnode] (good) at (6.8,-1.15) {\sudokuStateC};
            \node[faillabel] (fail) at (9.7,1.15) {constraint\\violated};
            \node[donelabel] (done) at (9.7,-1.15) {Done!};

            \draw[goodedge] (s0) -- node[steplabel, pos=0.5, yshift=7pt] {$r_1c_2=4$} (s1);
            \draw[badedge] (s1) -- node[steplabel, pos=0.45, yshift=9.5pt] {$r_4c_2=3$} (bad);
            \draw[goodedge] (s1) -- node[steplabel, pos=0.45, yshift=-9.5pt] {$r_4c_2=2$} (good);
            \draw[badedge] (bad) -- (fail);
            \draw[backtrackedge] (fail.south) to[out=-105,in=22] node[steplabel, pos=0.3, yshift=-10pt] {backtrack} (s1.east);
            \draw[goodedge] (good) -- (done);

            \node[draw=black!25, rounded corners=2pt, fit=(s0)(s1)(bad)(good)(fail), inner sep=6pt] {};
        \end{tikzpicture}%
    }
    \caption{Example of Sudoku $4{\times}4$ trees.}\label{fig:sudoku}
\end{figure*}
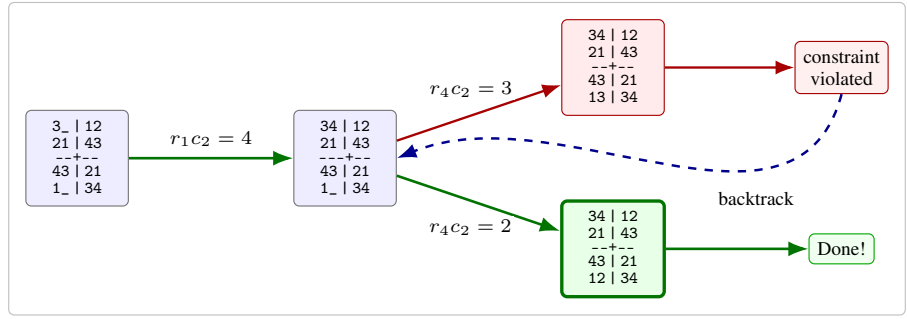

\begin{table}[ht]
    \centering
    \scriptsize
    \renewcommand{\arraystretch}{1.8}
    \begin{tabularx}{\linewidth}{
        >{\RaggedRight\arraybackslash}p{0.12\linewidth}
        >{\RaggedRight\arraybackslash}X
        >{\RaggedRight\arraybackslash}p{0.32\linewidth}}
        \toprule
        \textbf{Type}    & \textbf{Description}                                                                                                                    & \textbf{Example} \\
        \midrule
        \textbf{Invalid} & Target is missing, points to the current last node, or is emitted after \texttt{<done>}.
                         & \parbox[t]{\hsize}{{\ttfamily <node>2 wrong</node>\par <backtrack>2</backtrack>}}                                                                          \\[6pt]
        \textbf{Valid}   & Target exists and is earlier than the current node, but it is not confirmed as the best correction point.
                         & \parbox[t]{\hsize}{{\ttfamily <node>1 correct</node>\par <node>2 correct</node>\par <backtrack>1</backtrack>}}                                             \\[6pt]
        \textbf{Correct} & Chosen target is a correct state, while the previous node is not correct.
                         & \parbox[t]{\hsize}{{\ttfamily <node>1 correct</node>\par <node>2 correct</node>\par <node>3 wrong</node>\par <backtrack>1</backtrack>}}                    \\[6pt]
        \textbf{Perfect} & Target is correct, previous node is incorrect, and the next node after the target is not correct.
                         & \parbox[t]{\hsize}{{\ttfamily <node>1 correct</node>\par <node>2 wrong</node>\par <node>3 wrong</node>\par <backtrack>1</backtrack>}}                      \\
        \bottomrule
    \end{tabularx}
    \vspace{4pt}
    \caption{Backtrack categories used for correction-aware evaluation. The categories are nested in quality: perfect backtracks are also correct, correct backtracks are also valid, and valid backtracks are also non-invalid.}\label{tab:backtrack-categories}
\end{table}

\subsection{Explorer schedules}\label{apx:explorer-details}

\begin{figure*}[hb]
    \centering
    \resizebox{0.96\linewidth}{!}{%
        \begin{tikzpicture}[
                >=Latex,
                rootnode/.style={circle, draw=black!55, fill=white, minimum size=8mm, inner sep=1pt, font=\scriptsize},
                chainnode/.style={circle, draw=blue!55!black, fill=blue!7, minimum size=6mm, inner sep=1pt, font=\scriptsize},
                sweetnode/.style={circle, draw=orange!65!black, fill=orange!10, minimum size=6mm, inner sep=1pt, font=\scriptsize},
                branchnode/.style={circle, draw=green!45!black, fill=green!8, minimum size=6mm, inner sep=1pt, font=\scriptsize},
                edge/.style={->, line width=0.75pt, draw=black!55},
                panel/.style={draw=black!35, fill=black!3, rounded corners=2pt, inner sep=4pt, font=\scriptsize, align=left}
            ]
            \node[rootnode] (lr) at (0,1.35) {Root};
            \node[chainnode] (l1) at (-1.35,0.15) {};
            \node[chainnode] (l2) at (0,0.15) {};
            \node[chainnode] (l3) at (1.35,0.15) {};
            \node[chainnode] (l11) at (-1.35,-0.95) {};
            \node[chainnode] (l22) at (0,-0.95) {};
            \node[chainnode] (l33) at (1.35,-0.95) {};
            \foreach \a/\b in {lr/l1,lr/l2,lr/l3,l1/l11,l2/l22,l3/l33}{\draw[edge] (\a) -- (\b);}
            \node[panel, text width=3.5cm] at (0,-2.05) {Linear: fixed-width parallel chains.};
            \node[draw=black!25, rounded corners=2pt, fit=(lr)(l1)(l2)(l3)(l11)(l22)(l33), inner sep=5pt] {};

            \node[rootnode] (sr) at (4.55,1.35) {Root};
            \node[sweetnode] (s1) at (3.3,0.15) {};
            \node[sweetnode] (s2) at (4.55,0.15) {};
            \node[sweetnode] (s3) at (5.8,0.15) {};
            \node[sweetnode] (s11) at (3.3,-0.95) {};
            \node[sweetnode] (s21) at (4.25,-0.95) {};
            \node[sweetnode] (s22) at (5.0,-0.95) {};
            \node[sweetnode] (s31) at (5.8,-0.95) {};
            \foreach \a/\b in {sr/s1,sr/s2,sr/s3,s1/s11,s2/s21,s2/s22,s3/s31}{\draw[edge] (\a) -- (\b);}
            \node[panel, text width=3.3cm] at (4.55,-2.05) {Sweet point: balances between linear and exponential};
            \node[draw=black!25, rounded corners=2pt, fit=(sr)(s1)(s2)(s3)(s11)(s21)(s22)(s31), inner sep=5pt] {};

            \node[rootnode] (er) at (9.2,1.35) {Root};
            \node[branchnode] (e1) at (7.95,0.15) {};
            \node[branchnode] (e2) at (10.45,0.15) {};
            \node[branchnode] (e11) at (7.35,-0.95) {};
            \node[branchnode] (e12) at (8.55,-0.95) {};
            \node[branchnode] (e21) at (9.85,-0.95) {};
            \node[branchnode] (e22) at (11.05,-0.95) {};
            \foreach \a/\b in {er/e1,er/e2,e1/e11,e1/e12,e2/e21,e2/e22}{\draw[edge] (\a) -- (\b);}
            \node[panel, text width=4.4cm] at (9.2,-2.05) {Exponential: produces same number of branches on each depth};
            \node[draw=black!25, rounded corners=2pt, fit=(er)(e1)(e2)(e11)(e12)(e21)(e22), inner sep=5pt] {};
        \end{tikzpicture}%
    }
    \caption{Exploration schedules. Linear exploration spawns a bounded number of parallel chains, while exponential exploration branches from each node.}\label{fig:explorer-strategies}
\end{figure*}
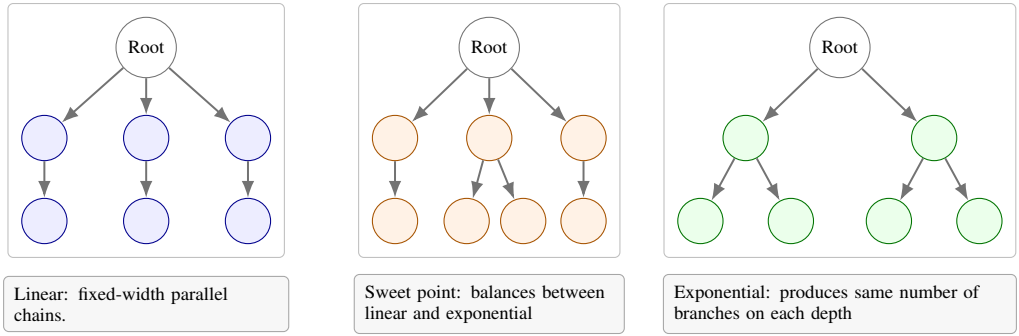

We study multiple explorers because the useful search regime is task-dependent. If the explorer branches too little, the model cannot recover from an early local mistake. If it branches too aggressively, the number of candidate partial chains grows too quickly and the exploration budget is spent on shallow alternatives rather than on finishing promising trajectories.

Let $B$ be the branching factor and let $d$ denote a frontier depth relative to the selected prefix root. A naive exponential branching policy allows every frontier to generate up to $B$ children,
\[
    c_{\exp}(d) = B,
\]
so the number of nodes reachable by depth $D$ grows on the order of
\[
    1 + B + B^2 + \cdots + B^D = O(B^D).
\]
This is undesirable in our setting because training requires validator calls and pair construction for explored branches; exponential growth therefore converts a small increase in depth into a large increase in cost while often failing to improve branch quality.

The linear explorer implements the opposite extreme. Its frontier capacity is
\[
    c_{\text{lin}}(d) =
    \begin{cases}
        B, & d = 0, \\
        1, & d > 0.
    \end{cases}
\]
This creates $B$ parallel chains rather than a dense tree: the root can spawn up to $B$ alternatives, and each alternative then continues as a single line. Linear exploration is attractive when depth matters more than local breadth, because cost remains predictable and the explorer can advance multiple candidate chains in parallel.

The dendritic explorer is designed for the intermediate regime. It keeps a bounded per-iteration child budget $U$ but scores frontiers so that search behavior shifts from breadth-first to depth-first as the local tree grows. Let $N$ be the number of discovered nodes in the current prefix-local tree. The transition is controlled by
\[
    \sigma(N) = \frac{1}{1 + \exp\left(-h\left(\log_B N - s\right)\right)},
\]
\[
    \alpha(N) = \alpha_{\text{bfs}} + \left(\alpha_{\text{dfs}} - \alpha_{\text{bfs}}\right)\sigma(N),
\]
where $s$ is the transition point and $h$ controls sharpness. Early in exploration, $\alpha(N)$ is close to $\alpha_{\text{bfs}}$ and the explorer favors shallower frontiers; later it approaches $\alpha_{\text{dfs}}$ and increasingly favors deeper continuations.

For an expandable frontier $i$, let $d_i$ be its relative depth, let $c_{\max}(i)$ be its maximum number of children, and let $\mathrm{gap}_i = c_{\max}(i) - \mathrm{children}_i$ denote remaining child capacity. The dendritic explorer uses the log-score
\[
    \log \mathrm{score}_i
    =
    \alpha(N)\log\!\left(1 + \frac{d_i}{D_i^{\max}}\right)
    +
    \beta\log\!\left(1 + \frac{\mathrm{gap}_i}{\max(1, c_{\max}(i))} + \varepsilon\right),
\]
where $D_i^{\max}$ is the remaining allowed depth from the current prefix and $\beta$ weights the unused-capacity term. After scoring frontiers, the explorer deterministically allocates the budget $U$ across the highest-scoring nodes. In practice, this yields a broad-then-deep pattern: the explorer first discovers competing branches and then concentrates resources on extending the most promising ones.

\subsection{SFT-B pair construction}\label{apx:sftb-pairs}

\begin{figure}[t]
    \centering

    \begin{subfigure}[t]{0.25\columnwidth}
        \centering
        \begin{lstlisting}[basicstyle=\ttfamily\scriptsize,columns=fullflexible]
Input:
[ <node>0 puzzle input</node>
  <node>1 Step A</node>
  <node>2 Step E</node> ]

Target:
<done>Solution!</done>
        \end{lstlisting}

        \caption{Success pair.}\label{fig:pairs-kind-a}
    \end{subfigure}\hfill
    \begin{subfigure}[t]{0.25\columnwidth}
        \centering
        \begin{lstlisting}[basicstyle=\ttfamily\scriptsize,columns=fullflexible]
Input:
[ <node>0 puzzle input</node>
  <node>1 Step A</node>
  <node>2 Step B</node>
  <node>3 Step C</node>
  <node>4 Step D</node> ]

Target:
<backtrack>
    1 Constraints violated
</backtrack>
        \end{lstlisting}

        \caption{Failed pair.}\label{fig:pairs-kind-b}
    \end{subfigure}\hfill
    \begin{subfigure}[t]{0.35\columnwidth}
        \centering
        \begin{lstlisting}[basicstyle=\ttfamily\scriptsize,columns=fullflexible]
Input:
[ <node>0 puzzle input</node>
  <node>1 Step A</node>
  <trace>
  Reason: Constraints violated
  Explored node: <node>2 Step B</node>
  Error node: <node>4 Step D</node>
  </trace> ]

Target:
<node>2 Step E</node>
        \end{lstlisting}
        \caption{Continue pair.}\label{fig:pairs-kind-c}
    \end{subfigure}

    \caption{Example of different SFT-B pair kinds for prefix \texttt{<node>0 puzzle input</node>, <node>1 Step A</node>}.}\label{fig:sftb-pair-kinds}
\end{figure}

The SFT-B dataset contains three kinds of pairs (\Cref{fig:sftb-pair-kinds}). Success pairs train the model to continue along valid generated or gold branches. Failure pairs train the model to emit a \texttt{<backtrack>} action when a branch violates task constraints. Continue pairs train the model to resume from a valid prefix after a compact trace of the failed branch has been inserted. Trace construction is described separately in \Cref{apx:traces}.

We use a reverse curriculum over prefixes. Early training explores near-solution prefixes, where the remaining continuation is short and validation feedback is dense. Later stages move the starting prefix closer to the root, increasing the amount of search the solver must perform before reaching a valid \texttt{<done>} action. For a chain $c_t = (a_0, a_1, \ldots, a_t), |c_t| = t+1$, the curriculum starts with $T=t$, so the solver is asked to generate $a_t$ given the prefix $(a_0, a_1, \ldots, a_{t-1})$. After the SFT-A, exploration, and SFT-B passes, $T=t-1$ is set, so the new prefix becomes $(a_0, a_1, \ldots, a_{t-2})$. This process repeats until $T=1$ is reached, at which point the solver is asked to produce the whole chain based on the task description only.

\subsection{Trace format and traced recovery}\label{apx:traces}

During traced inference, when the model emits
\[
    \texttt{<backtrack>ID REASON</backtrack>},
\]
the runtime truncates the context to the target node and injects a compact trace:
\[
    \texttt{<trace>REASON SUMMARY</trace>}.
\]
The trace is never generated by the model as a target action. It is inserted by the runtime and appears only in the subsequent context. In our implementation the summary records the validator reason, the first explored action after the backtrack target, and the final erroneous action in the failed branch:
\begin{verbatim}
<trace>
    Reason: Invalid step: must fill exactly one empty cell
    Explored node: <node>2 ...</node>
    Error node: <node>4 ...</node>
</trace>
\end{verbatim}

Introducing traces requires careful handling to avoid distribution shift. The ordinary SFT-B failure pair teaches the model to emit a backtrack, but does not by itself teach the model how to continue after the runtime inserts a trace. To mitigate this mismatch, we create synthetic traces during SFT-B training: for each failure pair, we insert a synthetic trace after the backtrack target and create a continuation target from the corresponding valid branch.

We evaluate three recovery regimes. In \emph{reset} mode, the runtime truncates the context to the target node and discards the failed branch. In \emph{preserve} mode, the raw \texttt{<backtrack>} action remains in context. In \emph{traced} mode, the runtime truncates to the target node and appends \texttt{<trace>...</trace>}. The Sudoku ablations in \Cref{fig:sudoku-results} compare traced and non-traced training, while the main Sudoku results in \Cref{tab:sudoku-results-summary} use traced recovery.

\subsection{Finetune expanders}\label{apx:finetune-expanders}

\begin{figure}[ht]
    \centering

    \begin{subfigure}[t]{0.33\columnwidth}
        \centering
        \begin{lstlisting}[basicstyle=\ttfamily\scriptsize,columns=fullflexible]
1. [<node>0 puzzle input</node>]
    --> <node>1 Step A</node>

2. [<node>0 puzzle input</node>
    <node>1 Step A</node>]
    --> <node>2 Step E</node>

3. [<node>0 puzzle input</node>
    <node>1 Step A</node>
    <node>2 Step E</node>]
    --> <done>Solution!</done>
        \end{lstlisting}

        \caption{\texttt{all} expander.}
    \end{subfigure}\hfill
    \begin{subfigure}[t]{0.33\columnwidth}
        \centering
        \begin{lstlisting}[basicstyle=\ttfamily\scriptsize,columns=fullflexible]
1. [<node>0 puzzle input</node>
    <node>1 Step A</node>]
    --> <node>2 Step E</node>

2. [<node>0 puzzle input</node>
    <node>1 Step A</node>
    <node>2 Step E</node>]
    --> <done>Solution!</done>
        \end{lstlisting}

        \caption{\texttt{prefix} expander.}
    \end{subfigure}\hfill
    \begin{subfigure}[t]{0.33\columnwidth}
        \centering
        \begin{lstlisting}[basicstyle=\ttfamily\scriptsize,columns=fullflexible]
1. [<node>0 puzzle input</node>
    <node>1 Step A</node>
    <node>2 Step B</node>
    <node>3 Step C</node>
    <node>4 Step D</node>]
    --> <backtrack>
            1 Constraints violated
        </backtrack>
        \end{lstlisting}
        \caption{\texttt{final} expander.}
    \end{subfigure}

    \caption{Finetune expander examples for prefix \texttt{<node>0 puzzle input</node>, <node>1 Step A</node>}. Format is \texttt{[training prefix] --> training target}.}\label{fig:finetune-expanders}
\end{figure}

Finetune expanders control which prefix-target pairs are materialized from a successful or failed chain before SFT-B (\Cref{fig:finetune-expanders}). We need multiple expanders because the same explored tree can emphasize different learning signals. The \texttt{all} expander supervises every intermediate continuation, producing dense next-step imitation. The \texttt{prefix} expander keeps later generated prefixes and the terminal answer, reducing dataset size while still teaching continuation from non-root contexts. The \texttt{final} expander is used for failed leaves: it trains the model to emit the final recovery action from the full failed branch. In Sudoku $4{\times}4$, the main traced setting combines prefix-style supervision for successful branches with final failed-branch supervision, so the model sees both post-trace continuation targets and explicit backtrack targets without over-weighting every intermediate failed prefix.

\subsection{Additional task methodology details}\label{apx:task-methodology-details}

\paragraph{Training and evaluation hyperparameters.}
Unless stated otherwise, most experiments use the following parameters. We train Qwen3-4B with the \texttt{max\_seq\_len}=2560, vLLM GPU memory utilization 0.8, gradient checkpointing and 10 dataloader workers and exploration batch size 256. For exploration we use a linear explorer with branching budget $B=5$, leaf multiplier $c_{\mathrm{leaf}}=3$, maximum leaf capability 20, adaptive $T_\text{max}$ (so the maximum curriculum $T$ is taken for each gold chain individually based on its length) and deduplication (what means we do not include exactly the same generated nodes on one depth into separate chains). We also use \emph{restricted solver} --- vLLM with restricted regex output on during diligent pipeline --- in order to generate well-structured chains on exploration phase. The SFT schedule is one epoch for SFT-A, one epoch for all SFT-B examples. The optimizer and adapter settings are QLoRA in bfloat16, learning rate $5\cdot10^{-5}$, LoRA rank 32 and alpha 64, per-device batch size 16, gradient accumulation 1, cosine learning-rate schedule, warmup ratio 0.03, weight decay 0.01, and LoRA dropout 0.05. For FT and Gold models we fine-tune for 3 epochs. Evaluation uses each task's maximum evaluation step budget: 30 steps for Sudoku $4{\times}4$, 30 for Sudoku w/ RT $4{\times}4$, 50 for Blocksworld and 30 for Hidden graph. We conducted experiment on NVIDIA A100 80Gb.

\paragraph{Hidden directed graph validation.}
For \texttt{<node>} actions, the hidden graph validator checks that the requested target is visible from the current node, rejects moves to \texttt{FAIL}, rejects moving to \texttt{GOAL} instead of emitting \texttt{<done>}, and requires \texttt{<done>} when \texttt{GOAL} is already visible. For terminal actions, it verifies that the answer is exactly the current path extended by \texttt{GOAL}, that \texttt{GOAL} is currently visible, and that \texttt{FAIL} is absent. A valid move can still reveal a losing branch: if the new observation is \texttt{X -> FAIL} or \texttt{X -> (none)}, the explorer records a failed leaf with a reason such as \texttt{Failure reached after START -> X}.

\paragraph{Sudoku validation.}
Every Sudoku action contains a complete grid rather than a coordinate-level edit. The solver is instructed to output exactly one tag per turn. If the current grid still contains empty cells, it may output a \texttt{<node>} that fills exactly one empty cell or a \texttt{<backtrack>} if the branch is invalid. If the current grid is already complete, it must output \texttt{<done>} with the same complete grid; filling the final empty cell and terminating are therefore two separate turns:

\noindent
\begin{minipage}[t]{0.25\linewidth}
    \begin{verbatim}
<node>0
3 4 | 1 2
2 1 | 4 3
----+----
4 3 | 2 1
1 _ | 3 4
</node>
\end{verbatim}
\end{minipage}\hfill
\begin{minipage}[t]{0.25\linewidth}
    \begin{verbatim}
<node>1
3 4 | 1 2
2 1 | 4 3
----+----
4 3 | 2 1
1 2 | 3 4
</node>
\end{verbatim}
\end{minipage}\hfill
\begin{minipage}[t]{0.5\linewidth}
    \begin{verbatim}
<done>
3 4 | 1 2
2 1 | 4 3
----+----
4 3 | 2 1
1 2 | 3 4
</done>
\end{verbatim}
\end{minipage}

The Sudoku validator preserves all initial clues, enforces row, column, and box constraints, and requires each \texttt{<node>} transition to place exactly one digit into an empty cell without clearing or rewriting any filled cell. For \texttt{<done>}, the validator additionally requires that the action repeats the latest complete grid and that no empty cells remain. During exploration, failures such as changing an initial clue, placing multiple digits at once, or emitting \texttt{<done>} before completion become validator reasons for backtrack targets.

\paragraph{Sudoku w/ RT generation and validation.}
Sudoku w/ RT supports grids from $4{\times}4$ to $9{\times}9$. Each chain begins with a partially filled Sudoku grid; every \texttt{<node>} action contains the updated grid followed by a brief justification separated by a blank line, and \texttt{<done>} must reproduce the fully solved grid. The primary training dataset contains 1{,}200 $4{\times}4$ puzzles with reasoning generated by Qwen3-32B. Grid transitions are validated using the standard Sudoku validator above, but the reasoning text is not validated.

\paragraph{Blocksworld generation and validation.}
Blocksworld initial and goal states are sampled uniformly at random with the \textsl{bwstates} utility and formatted as PDDL problems under the standard blocksworld-4ops domain. Duplicate problems and already-solved initial states are discarded; generation moves to the next block count when more than 90\% of samples in a rolling window of 2048 tasks are duplicates. Valid problems are solved with Fast-Forward (FF), and the returned plans are converted from PDDL into text prompts with color names for blocks. VAL validates candidate outputs by simulating plan execution from the initial state, checking action preconditions and effects at each step, and confirming that the resulting state satisfies the goal.

\subsection{Sudoku w/ RT task reasoning generation prompt}
\label{apx:sudoku-extended-generation-prompt}
For each pair of consecutive grids $(G_t, G_{t+1})$ in a solution chain, the LLM receives a prompt containing the current grid and the next grid (with one cell filled) in visual format, optionally preceded by the reasoning from the previous step for context. The prompt instructs the model to identify which cell changed, state what digit was placed, and explain why that digit is the only valid option using standard Sudoku constraints. The model is asked to respond concisely in 1--3 sentences without numbering or meta-references.

The full prompt template is:

\begin{figure*}[t]
    \centering
    \includegraphics[width=.55\linewidth]{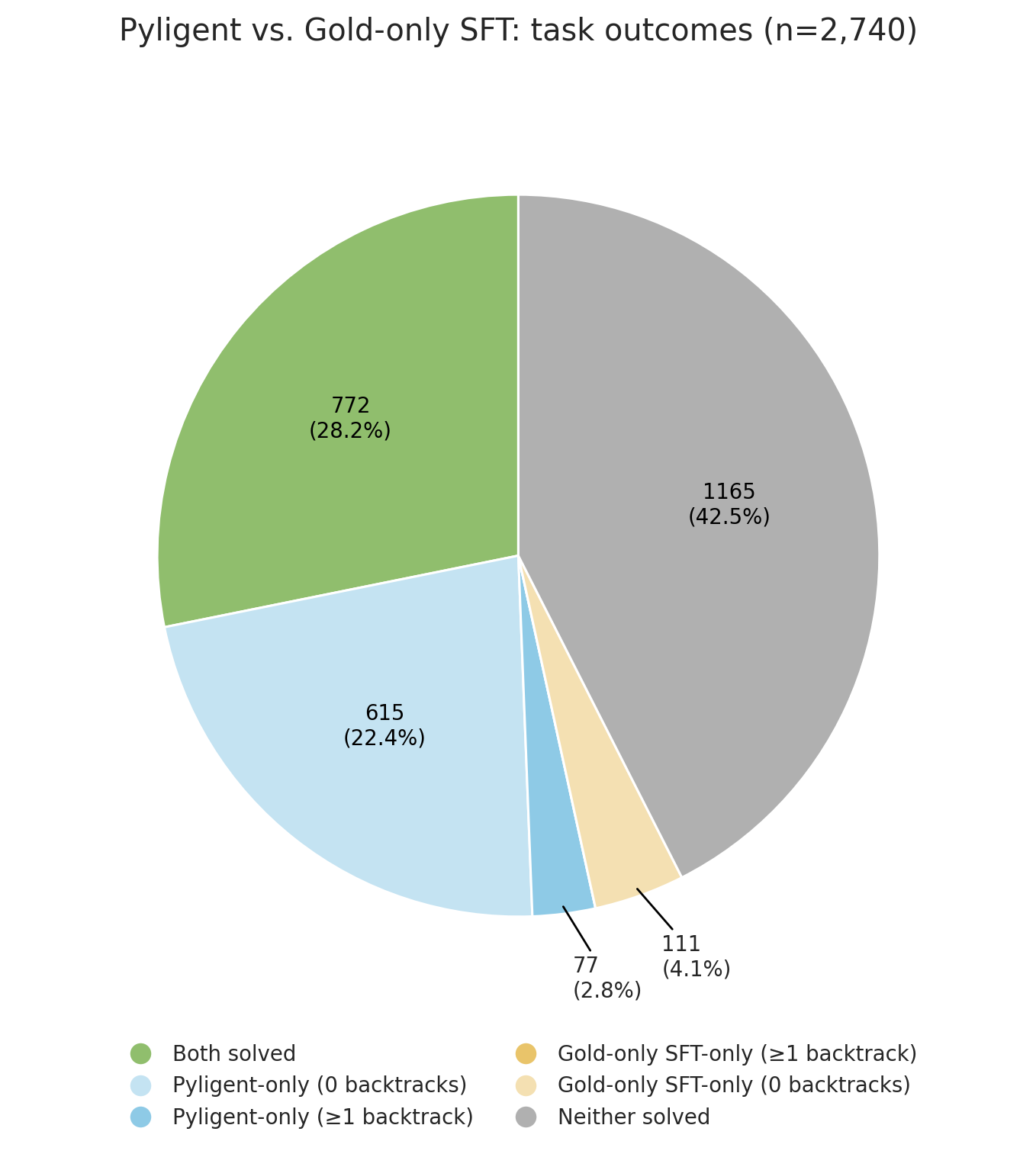}
    \caption{Distribution of task outcomes on the Blocksworld test set for the Gold-only SFT model and the Pyligent model, aggregated across 3 runs. "Both solved" indicates that both models reached a valid \texttt{<done>} node and completed the task successfully in all 3 runs. The remaining slices show which tasks were solved uniquely by one model or left unsolved by both.}\label{fig:blocksworld-outcomes}
\end{figure*}

\begin{verbatim}
You are analyzing a single-cell transition in a {size}x{size} Sudoku puzzle.

Current grid:
{current_visual}
[Previous reasoning (if available):
{previous_reasoning}]

Next grid (one cell was filled):
{next_visual}

Explain the logical reasoning that justifies the newly filled cell.
Use only standard Sudoku constraints (row, column, and box rules).

Important rules:
- Identify which cell changed and what digit was placed.
- Explain why this digit is the only valid option for that cell.
- Do NOT mention 'Next grid' or 'Current grid' in your reasoning.
- Write concisely in 1-3 sentences.

Answer directly with the reasoning. Do not number your response.
\end{verbatim}

\subsection{Blocksworld task success distribution analysis}
\label{apx:blocksworld-analysis}

\Cref{fig:blocksworld-outcomes} shows the solvability of each task across all 3 runs. Both models were able to solve 28.2\% of tasks and failed on 42.5 \% of tasks. The Gold-only SFT model never succeeded via backtracking and solved only 4.1\% of tasks uniquely. In comparison, the Pyligent model was able to complete 25.2\% of tasks uniquely, and more than 11\% of these successes were achieved via backtracking. This indicates that the Pyligent training endowed the model with backtracking capabilities and improved its overall task understanding.

\begin{figure*}[t]
    \centering
    \includegraphics[width=.75\textwidth]{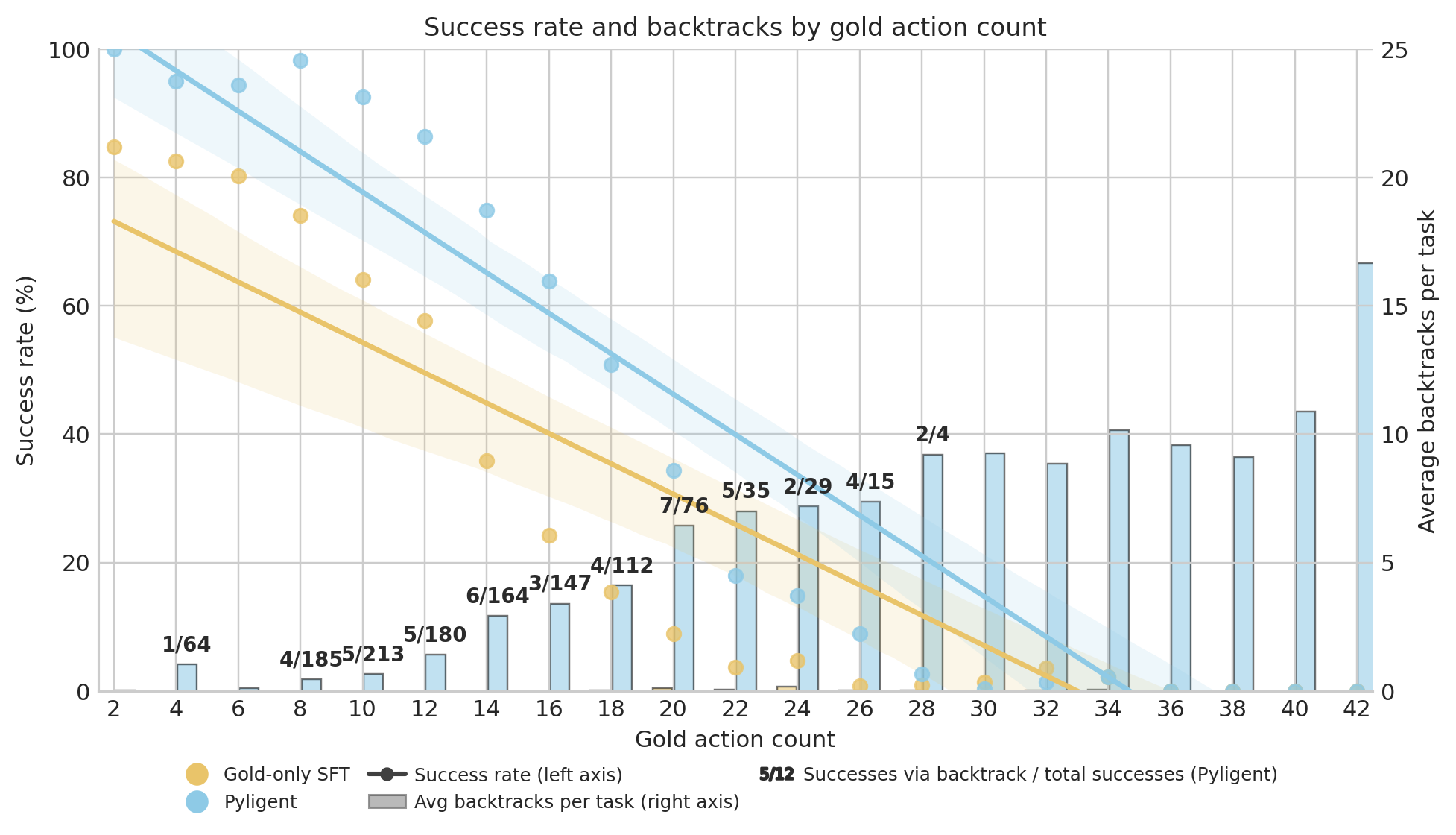}
    \caption{Distribution of Blocksworld successes and backtracks across the number of actions in the gold chains. The number of actions correlates with task difficulty, though the number of blocks may vary. The lines show the trend in success rate for Gold-only SFT model and Pyligent model, each bar shows the average number of backtracks per task, and the number above each bar indicates how many of the successful tasks involved at least one backtrack. The absence of a number indicates that none of the successes involved a backtrack.}\label{fig:blocksworld-action-distribution}
\end{figure*}

In \Cref{fig:blocksworld-action-distribution}, the success rate of both models declines as the number of actions in the gold plan increases, confirming that action count is a reasonable proxy for task difficulty. This decline also reflects the models' extrapolation capabilities, since both were trained only on tasks with up to 16 actions per gold chain. The Pyligent model consistently outperforms the Gold-only SFT model across the observed range, maintaining a higher success rate even as task complexity grows. The average number of backtracks per task for Pyligent also increases with action count, suggesting that the model relies on backtracking more heavily as tasks become harder. Moreover, the contribution of backtracking to overall success grows with task difficulty: at the hardest tasks, backtracking accounted for roughly 50\% of successes at 28 actions, albeit based on a small number of successful tasks, indicating that backtracking becomes increasingly important for solving the hardest tasks.

\end{document}